\documentclass[sigconf]{acmart}
\usepackage{acmart-taps,stfloats}
\AtBeginDocument{%
  }

\setcopyright{acmlicensed}
\copyrightyear{2025}
\acmYear{2025}
\acmDOI{10.1145/3715275.3732204}
\acmConference[FAccT '25]{ACM Conference on Fairness, Accountability and Transparency}{June 23--26,
  2025}{Athens, Greece}
\acmISBN{978-1-4503-XXXX-X/2018/06}

\usepackage[utf8]{inputenc} 
\usepackage[T1]{fontenc}
\usepackage{hyperref}   
\usepackage{url}
\usepackage{booktabs}   
\usepackage{amsfonts}   
\usepackage{nicefrac}   
\usepackage{microtype}  
\usepackage{xcolor} 

\usepackage{subcaption}
\usepackage{graphicx}
\usepackage{amsmath}
\usepackage{mathtools}
\usepackage{bbm}

\begin{document}

\title{Detecting Prefix Bias in LLM-based Reward Models}
\author{Ashwin Kumar}
\authornote{Corresponding Author}
\authornote{Work done as an intern at Meta Platforms, Inc.}
\affiliation{%
  \institution{Washington University in St Louis}
  \city{St Louis}
  \country{USA}
}
\email{ashwinkumar@wustl.edu}

\author{Yuzi He}
\authornotemark[1]
\affiliation{
\institution{Meta Platforms, Inc.}
  \city{Menlo Park}
\country{USA}
}
\email{yuzihe12@gmail.com}

\author{Aram H. Markosyan}
\affiliation{
\institution{Meta Platforms, Inc.}
  \city{Menlo Park}
\country{USA}
}
\email{aram.math@gmail.com}

\author{Bobbie Chern}
\affiliation{
\institution{Meta Platforms, Inc.}
  \city{Sunnyvale}
\country{USA}
}
\email{bgchern@meta.com}

\author{Imanol Arrieta-Ibarra}
\authornote{Work done while at Meta Platforms, Inc.}
\affiliation{
\institution{Independent}
  \city{San Mateo}
\country{USA}
}
\email{imanol.arrieta.ibarra@gmail.com}

\renewcommand{\shortauthors}{Kumar et al.}

\begin{abstract}

Reinforcement Learning with Human Feedback (RLHF) has emerged as a key paradigm for task-specific fine-tuning of language models using human preference data. While numerous publicly available preference datasets provide pairwise comparisons of responses, the potential for biases in the resulting reward models remains underexplored. In this work, we introduce novel methods to detect and evaluate prefix bias—a systematic shift in model preferences triggered by minor variations in query prefixes—in LLM-based reward models trained on such datasets.  We leverage these metrics to reveal significant biases in preference models across racial and gender dimensions. Our comprehensive evaluation spans diverse open-source preference datasets and reward model architectures, demonstrating susceptibility to this kind of bias regardless of the underlying model architecture. Furthermore, we propose a data augmentation strategy to mitigate these biases, showing its effectiveness in reducing the impact of prefix bias. Our findings highlight the critical need for bias-aware dataset design and evaluation in developing fair and reliable reward models, contributing to the broader discourse on fairness in AI.
\end{abstract}





\maketitle
\section{Introduction}
Large Language Models (LLMs) have seen a large growth in research and usage~\cite{gpt4, llama2}. Their ability to understand context and generate and analyze text has proven their utility in applications from code completion~\cite{codeLlama} to medical diagnosis~\cite{MedPalm2}.  This is thanks to the large amounts of pre-training that models undergo, which deployers can leverage for different use cases. However, to efficiently use LLMs for different tasks, we need to fine-tune them on task-specific information~\cite{AnthropicHH, InstructGPT, learningToSummarize}. 

This finetuning can be done in multiple ways, the simplest being Supervised Fine Tuning (SFT), where domain data is used to train the model with next-token prediction. Reinforcement Learning with Human Feedback (RLHF) is a paradigm popularized by the InstructGPT paper~\cite{InstructGPT}, and further used by Anthropic~\cite{AnthropicHH}, Llama~\cite{llama2, llama} and other popular language models. When using RLHF, a preference model or Reward Model (RM) is trained to discriminate between pairwise responses to an input prompt to give higher scores for responses that are \emph{preferred} by human reviewers~\cite{InstructGPT}. These reward models are used for reinforcement learning to guide fine-tuning, using algorithms like proximal policy optimization (PPO)~\cite{schulman2017PPO}.

There have been studies~\cite{LLMBiasSurvey, societalbiasLLms, stochasticParrots} looking at bias in language models. However, to the best of our knowledge, no one has studied bias in reward models in the context of RLHF fine-tuning. Further, there has been limited study of \emph{prefix bias}, where changing the prefix to a prompt can change the model's behavior.
RMs are typically used to evaluate prompt-response pairs and assign a value to them, which then trains the finetuned model to prefer the response that receives a higher reward. If the reward model’s preferences are biased, it may poison the downstream model and cause undesirable learned behavior. Thus, it is important to create ways to identify and mitigate bias in the reward model stage.

In this paper, we study bias in reward models by explicitly conditioning candidate responses on different demographic identifiers using short textual prefixes. Our contributions are as follows:
\begin{enumerate}
\item We design \textbf{auto-influence} and \textbf{cross-influence}, two methods for evaluating the prefix-bias in a learned reward model. We also present \textbf{winrate deviation} and \textbf{accuracy deviation}, metrics for quantifying auto-influence and cross -influence respectively.
\item We train and evaluate reward models on a variety of open datasets and show the presence of racial and gender bias using these metrics. We also use different model architectures and show that this bias is present for all of them, suggesting that the bias is learned from the datasets.
\item We conduct experiments to identify the source of the bias and uncover patterns based on the training dataset used, and present a solution using data augmentation to fix this issue.
\end{enumerate}

\section{Related Work}
\paragraph{Bias in LLMs:}
The study of bias in LLMs has become a pressing and active field of research~\cite{LLMGenderbias, LLMBiasSurvey, LLMraceBiasName, racistMedicalLLMs}, driven by their widespread use in sensitive domains. Early work focused on how training data can encode systemic biases~\cite{WebCorpusBias, stochasticParrots}, while more recent research has highlighted specific downstream harms, such as racial disparities in clinical advice~\cite{racistMedicalLLMs} and gendered patterns in name-based completions~\cite{LLMraceBiasName, LLMGenderbias}. A recent survey~\cite{LLMBiasSurvey} provides a broad overview of these challenges. 
Prefix bias has also been studied in recent work; for instance, \citet{chaudhary2024prefixbiascert} use prefix perturbations to measure LLM robustness in a certification setting, but they do not consider the reward modeling stage. 

Several recent studies have taken an identity-grounded approach to evaluating bias in LLMs and reward models. \citet{kantharuban2024stereotype} find that chatbots often reflect racial stereotypes based on inferred user identity, blurring the line between personalization and harm. \citet{eloundou2024firstperson} introduce the concept of \emph{first-person fairness}, showing how name-based probes can surface demographic biases in chatbot behavior. These works underscore the importance of evaluating LLM behavior through demographic context. 
However, in this paper, we focus on developing metrics to evaluate and quantify prefix bias in Reward Models.
We explore the biases that may exist and appear in the Reward Model stage, which can introduce biases to the fine-tuned model due to an inaccurate learned reward function. As such, this paper is orthogonal to the rich field of analyzing bias in pre-trained and deployed LLMs.

\paragraph{Language Model Fine-tuning:}
Trained on vast corpora, language models also possess vast information. However, to constrain their outputs to suit a particular use case, and to ensure their outputs are aligned with our expectations, LLMs need to be fine-tuned on curated data~\cite{AnthropicHH, InstructGPT, learningToSummarize}. Reinforcement Learning from Human Feedback (RLHF)~\cite{InstructGPT, AnthropicHH} is a popular technique for achieving LLM alignment. RLHF aims to collect human preferences in terms of pairwise rankings, which are then used to learn a preference model or Reward Model (RM). The RM is then used to fine-tune the LLM using Proximal Policy Optimization (PPO)~\cite{schulman2017PPO}, which is a popular reinforcement learning algorithm. Reward models may also be used for best-of-n (BoN) sampling of the model outputs. While there exist other finetuning methods like supervised finetuning (SFT) and direct preference optimization (DPO)~\cite{rafailov2024DPO}, we focus on methods that rely on reward models, because of their widespread use.

\paragraph{Reward Models:}
It is known that reward models can have generalization and misspecification issues~\cite{RLHFIssues}. Recent work on reward model overoptimization~\cite{RMscaling} has shown that larger reward models exhibit lower susceptibility to reward hacking, and increasing training data reduces overoptimization. This study considered reward models with up to 3 Billion parameters. 
A follow-up paper showed that using ensembling in reward models helps to reduce overoptimization~\cite{RMEnsembles}. However, these papers do not consider any bias resulting from overoptimization. In a more related vein, \citet{mire2025rejected} demonstrated that reward models penalize African American Language (AAL), leading to representational disparities.

To the best of our knowledge, bias in LLM-based reward models using prefix-based attacks has not been studied, and our paper is the first to show its existence when training popular open-weight language models~\cite{llama, llama2, falcon, opt, flan} on popular RLHF datasets~\cite{AnthropicHH, SHP_paper}.

\section{Learning Reward Models}
Reward models are an integral part of reinforcement learning systems. They endow the models with the ability to reason about their actions and provide an optimization target. A reward model takes as input a combination of states and actions and tells the RL agent the associated reward. Since the objective of RL is typically to maximize the total or expected reward, agents learn to take actions with the best-expected returns given the current state. If the reward is misspecified, the policy may not reflect the intended behavior.
In the context of LLM fine-tuning, the RM is intended to provide feedback about the LLM's generated output, and RLHF aims to use the RM to make the model learn to produce output that has a higher reward.

The reward model is initialized using a pre-trained LLM with an added value head.
The RM then takes a sequence as input and predicts a scalar reward for it. 
To learn a reward model, the technique proposed by InstructGPT~\cite{InstructGPT} is to get human reviewers to rank a set of responses to an input based on the criteria of choice (e.g. helpfulness, harmlessness, instruction following). Then, using this ranking, we can learn a reward model by minimizing the loss in Equation~\ref{eq:loss_fn}, where $\sigma$ is the sigmoid function.

\begin{equation}
\mathcal{L} = -\log(\sigma(S(c) - S(r))) \label{eq:loss_fn}
\end{equation}

Here, $S(t)$ is the reward model's score for text $t$, and $c$ and $r$ are the chosen (preferred) response text and rejected (less preferred) response text respectively, based on the human reviewer's ranking. We describe how $c$ and $r$ are constructed in Section~\ref{sec:RMEval}.

\subsection{Using Reward Models for Evaluation}
\label{sec:RMEval}
In this section, we describe our preprocessing steps and evaluation methodology.
Reward models are used to compare different responses to the same query. Given a dataset $D$, we define an input data point as the tuple $D_i = \langle q, a_1, a_2 \rangle$ containing a query $q$ and two responses $a_1$ and $a_2$. Without loss of generality, we use the convention that $a_1$ is the preferred/better response. Many RLHF datasets~\cite{AnthropicHH, SHP_paper} datasets are available in this format. For training and evaluation using such datasets, we preprocess the dataset using the following template:

\begin{align}
T(q, a_1, a_2) = &\text{``Prompt:''} + q + \text{``Response1:''} + a_1 + \\
&\text{``Response2:''} + a_2 + \\
&\text{``Is response 1 better than response 2? A:''}
\end{align} 

This template, formatted as a natural language question, leverages the reward model's linguistic capabilities. However, the model's output ($S(t)$) remains a scalar value. Sequences $c$ (chosen) and $r$ (rejected) are constructed for each data point by flipping the order of responses to mitigate input-order bias:
\begin{align}
c &= T(q, a_1, a_2)\\
r &= T(q, a_2, a_1) 
\end{align}

Then, $S(c) - S(r)$ is the reward model's preference for $a_1$ versus $a_2$. We then compute the scores $c$ and $r$ and plug them into the loss (Eq.\ref{eq:loss_fn}) to train the reward model. This technique of explicitly asking the model to compare both responses results in a better agreement rate compared to directly scoring each response and can be adapted during PPO training by making $a_2$ be the empty string to get independent scores~\cite{SHP_paper}\footnote{A similar structure is used by SteamSHP, a reward model released by \citet{SHP_paper}}.

To measure how well a reward model performs, we use the \textbf{agreement rate}, which encodes how often the reward model's preferences align with the human preferences. We concisely represent the above process for a trained reward model $M$ as $M(q, a_1, a_2)$, a shorthand for preprocessing the sample, computing rewards for each input and comparing them (Eq.\ref{eq:evaluate_input}). This also serves as the accuracy function.
\begin{align}
M(q,a_1,a_2) = \mathbb{1}[S(c) > S(r)] \label{eq:evaluate_input}
\end{align}

\section{Evaluating Prefix Bias in Trained Reward Models}

Reward models are intended to assign higher scores to responses that better align with human preferences, ideally focusing on the semantic quality of those responses. In this work, we investigate whether reward models' preferences can systematically shift when the same response is framed with different demographic indicators—such as race or gender—even when the underlying content remains unchanged.

To test this, we introduce a controlled intervention: we prepend short identity-related prefixes (e.g., “I am a woman.” or “I am a man.”) to responses before passing them to the reward model. These prefixes serve as stand-ins for demographic context. They do not reflect how users typically phrase inputs or how RLHF data is collected, but allow us to isolate whether the reward model’s scoring function changes based solely on these surface-level cues.

While this setup uses explicit markers, similar forms of demographic information—whether mentioned earlier in a conversation, embedded in system prompts, or inferred from user profiles—can realistically appear in deployed systems. Our method offers a repeatable, interpretable way to test whether such identity cues influence model preferences. We refer to this effect as \textbf{\textit{prefix bias}}.

Prefix bias can have real-world implications. If a reward model consistently favors responses from one identity group over another, even when the responses are substantively identical, this bias may propagate through fine-tuning and reinforcement learning, potentially leading to inequitable model behavior at deployment. Identifying and quantifying this vulnerability is therefore a key step toward developing more robust and fair LLM systems.

To evaluate prefix bias in reward models, we ask two questions:

\hspace{1em}\textbf{Q1. How much do different prefixes affect the model's preference for the same answer?}

\hspace{1em}\textbf{Q2. How does the addition of prefixes affect model accuracy?}

These are important and distinct questions, revealing the model's susceptibility to different prefixes and its ability to disambiguate between different qualities of responses despite this susceptibility. 

Here, we describe \textbf{Auto-Influence} and \textbf{Cross-Influence}, our methods to quantify these effects. Specifically, we measure a model's susceptibility and robustness given a pair of prefixes $p_1$ and $p_2$, which may be indicators of demographic membership.

\subsection{Auto-Influence}
Auto-Influence measures the sensitivity of the reward model to different prefixes while keeping the other text constant. Specifically, for a dataset $D_u$ consisting of unique query-response pairs $q \rightarrow a$, we calculate the \emph{winrate} $w(D_u, M, p_1, p_2)$ of reward model $M$ as:
\begin{equation}
w(D_u, M, p_1, p_2) = \frac{1}{|D_u|}\sum_{q,a \in D_u} M(q, p_1 + a, p_2 + a)
\end{equation}
where $p_1$ and $p_2$ are the prefixes applied to the response $a$. This winrate represents the model's preference for $p_1$ over $p_2$. If the model is unbiased, the winrate should average to $0.5$. Deviations from this value indicate susceptibility to prefix bias. The auto-influence is then captured as the magnitude of the \emph{winrate deviation} $\omega$:

\begin{equation}
\fbox{$
\omega(D_u, M, p_1, p_2) = w(D_u, M, p_1, p_2) - 0.5
$
}
\end{equation}

\subsection{Cross-Influence}
Cross-influence assesses the model's robustness when different prefixes are applied to correct and incorrect answers. Unlike auto-influence, which examines preference shifts, cross-influence evaluates the model's ability to maintain accurate rankings despite the introduction of prefixes. Accuracy, a key component of this metric, is defined as:
\begin{equation}
acc(D, M, p_1, p_2) = \frac{1}{|D|} \sum_{q,a_1,a_2\in D} M(q,p_1+a_1, p_2+a_2)
\end{equation}
where $acc(D, M, p_1, p_2)$ measures the proportion of cases where the model correctly ranks $a_1$ above $a_2$ given prefixes $p_1$ and $p_2$.
Using this, the cross-influence is calculated as the magnitude of the \emph{accuracy deviation} $\alpha$:
\begin{equation}
\fbox{$
\alpha(D, M, p_1, p_2) = acc(D, M, p_1, p_2) - acc(D, M, p_e, p_e)
$
}
\end{equation}
where $p_e$ denotes an empty prefix. Here, $\alpha(D, M, p_1, p_2)$ quantifies how much the accuracy changes when prefixes are introduced. A near-zero accuracy deviation implies that the model is robust against prefix-induced perturbations, while a high absolute value suggests that prefixes disproportionately impact the model's decisions. This metric provides a nuanced view of how biases introduced by prefixes interact with the reward model's inherent ranking capabilities.\footnote{These metrics are also laid out in Table \ref{tab:notation_reference} in the Appendix.}

\bigskip
These two metrics can yield complementary insights. For instance, a model may exhibit high auto-influence (strong preference for one prefix) but low cross-influence (overall accuracy remains unaffected), suggesting that prefix bias does not overwhelm response quality. Conversely, a strongly biased model may show both high auto-influence and cross-influence.

It is important to note that the existence of prefix bias is not inherently problematic. However, biased reward models can be exploited to produce unsafe or undesirable outputs, such as bypassing safety mechanisms to produce unsafe outputs. Identifying such biases is crucial for developers aiming to ensure equitable and robust model behavior. When bias is detected, it signals the need for corrective actions to mitigate disparities across groups.

\aptLtoX[graphic=no,type=html]{\begin{table}[]
\caption{Gender group prefixes used.}
\centering
\begin{tabular}{|c|c|} \hline 
 Name& Prefix\\ \hline 
 $P_e$& ""\\ \hline 
 $P_{wo}$& "I am a woman. " \\ \hline 
 $P_m$  & "I am a man. "\\ \hline
\end{tabular}
\label{tab:gender_prefixes}
\end{table}
\begin{table}
\centering
\caption{Race group prefixes used.}
\begin{tabular}{|c|c|} \hline 
 Name& Prefix\\ \hline 
 $P_e$& ""\\ \hline 
 $P_b$ & "I am black." \\ \hline
 $P_w$ & "I am white." \\ \hline
 $P_h$ & "I am hispanic." \\\hline
\end{tabular}
\label{tab:race_prefixes}
\end{table}
}{
\begin{table}[]
\begin{minipage}{0.45\textwidth}
\caption{Gender group prefixes used.}
\centering
\begin{tabular}{|c|c|} \hline 
 Name& Prefix\\ \hline 
 $P_e$& ""\\ \hline 
 $P_{wo}$& "I am a woman. " \\ \hline 
 $P_m$  & "I am a man. "\\ \hline
\end{tabular}
\label{tab:gender_prefixes}
\end{minipage}
\begin{minipage}{0.45\textwidth}
\centering
\caption{Race group prefixes used.}
\begin{tabular}{|c|c|} \hline 
 Name& Prefix\\ \hline 
 $P_e$& ""\\ \hline 
 $P_b$ & "I am black." \\ \hline
 $P_w$ & "I am white." \\ \hline
 $P_h$ & "I am hispanic." \\\hline
\end{tabular}
\label{tab:race_prefixes}
\end{minipage}
\end{table}}
\section{Experiments}

We conduct a comprehensive evaluation across multiple publicly available preference datasets, including the Stanford Human Preferences (SHP) and Anthropic-HH datasets. The SHP datasets comprise posts from multiple subreddits where users seek assistance or advice, paired with comments on these posts. Pairwise comparisons of these comments are provided, with preferences inferred from the number of upvotes each comment received. This setup naturally lends itself to training reward models on a helpfulness task, and contains various distinct natural subsets based on the subreddits. Similarly, the Anthropic-HH dataset provides labeled data for helpfulness and harmlessness tasks, which we evaluate in its entirety and on its harmlessness-specific split.

Our experimental process follows a structured methodology. First, we select a model architecture (e.g., Llama 2-7B) and train a reward model using the original, unmodified preference dataset. Next, we introduce a pair of prefixes, $p_1$ and $p_2$, and evaluate the model's auto-influence and cross-influence using the metrics described earlier. This process is repeated for all prefix pairs within a predefined set, $p_1, p_2 \in P$. The choice of prefixes $P$ is task-dependent; for example, we use prefixes relevant to detecting gender and racial biases in our experiments (Tables~\ref{tab:gender_prefixes} and~\ref{tab:race_prefixes}).
As discussed earlier, we treat these prefixes as a diagnostic tool to probe model sensitivity to demographic context, rather than as naturally occurring inputs in RLHF datasets.

For each evaluation, we construct a matrix of pairwise comparisons. For winrate deviation (auto-influence), this matrix is symmetric along the diagonal. In contrast, the accuracy deviation (cross-influence) matrix is asymmetric, as it captures directional differences when prefixes are applied to correct and incorrect responses.

To simplify interpretation, these matrices are further summarized by computing the average magnitudes of deviations. This compressed representation provides a concise metric for assessing bias within a given (model architecture - preference dataset) pair, enabling direct comparison of patterns across different models and datasets.

\aptLtoX[graphic=no,type=html]{\begin{table}
\centering
\caption{Winrate deviation for Llama 2 7B, legaladvice dataset (gender)}
\begin{tabular}{|c|c|c|c|}
\hline
$p_1$& \multicolumn{3}{|c|}{$p_2$}\\ \hline
 &  $P_e$&  $P_m$& $P_{wo}$\\ \hline
 $P_e$&  -&  -0.4297& -0.4884\\ \hline
 $P_m$&  0.4297&  -& -0.4046\\ \hline
 $P_{wo}$&  0.4884&  0.4046& -\\ \hline
\end{tabular}
\label{tab:gend_legal_winrate_dev}
\end{table}
\begin{table}
\centering
\caption{Accuracy deviation (percentage) for Llama 2 7B, legaladvice dataset (gender)}
\begin{tabular}{|c|c|c|c|}
\hline
$p_1$& \multicolumn{3}{|c|}{$p_2$}\\\hline
 &  $P_e$&  $P_m$& $P_{wo}$\\ \hline
 $P_e$&  0\%&  -3.66\%& -17.96\%\\ \hline
 $P_m$&  1.62\%&  -0.37\%& -9.16\%\\ \hline
 $P_{wo}$&  8.95\%&  5.81\%& -0.1\%\\ \hline
\end{tabular}
\label{tab:gend_legal_acc_dev}
\end{table}}{\begin{table}
\begin{minipage}{0.40\textwidth}
\centering
\caption{Winrate deviation for Llama 2 7B, legaladvice dataset (gender)}
\begin{tabular}{|c|c|c|c|}
\hline
$p_1$& \multicolumn{3}{|c|}{$p_2$}\\ \hline
 &  $P_e$&  $P_m$& $P_{wo}$\\ \hline
 $P_e$&  -&  -0.4297& -0.4884\\ \hline
 $P_m$&  0.4297&  -& -0.4046\\ \hline
 $P_{wo}$&  0.4884&  0.4046& -\\ \hline
\end{tabular}
\label{tab:gend_legal_winrate_dev}
\end{minipage}
\hspace{1em}
\begin{minipage}{0.40\textwidth}
\centering
\caption{Accuracy deviation (percentage) for Llama 2 7B, legaladvice dataset (gender)}
\begin{tabular}{|c|c|c|c|}
\hline
$p_1$& \multicolumn{3}{|c|}{$p_2$}\\\hline
 &  $P_e$&  $P_m$& $P_{wo}$\\ \hline
 $P_e$&  0\%&  -3.66\%& -17.96\%\\ \hline
 $P_m$&  1.62\%&  -0.37\%& -9.16\%\\ \hline
 $P_{wo}$&  8.95\%&  5.81\%& -0.1\%\\ \hline
\end{tabular}
\label{tab:gend_legal_acc_dev}
\end{minipage}
\end{table}}

\aptLtoX[graphic=no,type=html]{\begin{table}[]
\centering
\caption{Winrate deviation for Llama 2 7B, legaladvice dataset (race)}
\begin{tabular}{|c|c|c|c|c|}
\hline
$p_1$& \multicolumn{4}{|c|}{$p_2$}\\ \hline
 &  $P_e$&  $P_b$& $P_h$&$P_w$\\ \hline
 $P_e$&  -&  -0.4942& -0.4471&-0.4059\\ \hline
 $P_b$&  0.4942&  -& 0.3189&0.2938\\ \hline
 $P_h$&  0.4471&  -0.3189& -&0.2338\\ \hline
$P_w$& 0.4059& -0.2938& -0.2338&-\\\hline
\end{tabular}
\label{tab:race_legal_winrate_dev}
\end{table}
\begin{table}
\centering
\caption{Accuracy deviation (percentage) for Llama 2 7B, legaladvice dataset (race)}
\begin{tabular}{|c|c|c|c|c|}
\hline
$p_1$& \multicolumn{4}{|c|}{$p_2$}\\ \hline
 &  $P_e$&  $P_b$& $P_h$&$P_w$\\ \hline
 $P_e$&  0\%&  -15.18\%& -11.62\%&-4.71\%\\ \hline
 $P_b$&  7.96\%&  -0.42\%& -1.68\%&7.17\%\\ \hline
 $P_h$&  7.33\%&  -1.94\%& 0.26\%&7.28\%\\ \hline
$P_w$& 0.84\%& -10.21\%& -9.21\%&0.42\%\\\hline
\end{tabular}
\label{tab:race_legal_acc_dev}
\end{table}}{\begin{table}[]
\begin{minipage}{0.4\textwidth}
\centering
\caption{Winrate deviation for Llama 2 7B, legaladvice dataset (race)}
\begin{tabular}{|c|c|c|c|c|}
\hline
$p_1$& \multicolumn{4}{|c|}{$p_2$}\\ \hline
 &  $P_e$&  $P_b$& $P_h$&$P_w$\\ \hline
 $P_e$&  -&  -0.4942& -0.4471&-0.4059\\ \hline
 $P_b$&  0.4942&  -& 0.3189&0.2938\\ \hline
 $P_h$&  0.4471&  -0.3189& -&0.2338\\ \hline
$P_w$& 0.4059& -0.2938& -0.2338&-\\\hline
\end{tabular}
\label{tab:race_legal_winrate_dev}
\end{minipage}
\hspace{1em}
\begin{minipage}{0.40\textwidth}
\centering
\caption{Accuracy deviation (percentage) for Llama 2 7B, legaladvice dataset (race)}
\begin{tabular}{|c|c|c|c|c|}
\hline
$p_1$& \multicolumn{4}{|c|}{$p_2$}\\ \hline
 &  $P_e$&  $P_b$& $P_h$&$P_w$\\ \hline
 $P_e$&  0\%&  -15.18\%& -11.62\%&-4.71\%\\ \hline
 $P_b$&  7.96\%&  -0.42\%& -1.68\%&7.17\%\\ \hline
 $P_h$&  7.33\%&  -1.94\%& 0.26\%&7.28\%\\ \hline
$P_w$& 0.84\%& -10.21\%& -9.21\%&0.42\%\\\hline
\end{tabular}
\label{tab:race_legal_acc_dev}
\end{minipage}
\end{table}}

\section{Results}

We evaluate multiple datasets and reward model architectures to analyze bias and robustness.
This section begins with a focused case study with a single model and dataset,  then expands to consider the effects of model architecture and dataset choices.

\begin{table*}
    \centering
    \caption{Average Winrate and accuracy deviations for different language model architectures for the SHP/legaladvice dataset. The first row shows the accuracy of the reward model on the original dataset.}
    \begin{tabular}{|l|p{1cm}|p{1cm}|p{1cm}|p{1cm}|p{1cm}|p{1cm}|p{1cm}|p{1cm}|} \hline 
          \textbf{Group} & \textbf{Metric} & \textbf{opt-350m} & \textbf{flan-t5-large} & \textbf{gpt-j-6b} & \textbf{falcon-7b} & \textbf{llama-7b} & \textbf{llama-2-7b} & \textbf{llama-2-13b} \\ \hline 
          & accuracy & 70.55\% & 79.48\% & 79.45\% & 79.35\% & 79.11\% & 79.06\% & \textbf{81.88\%} \\ \hline 
  gender & $\bar{\omega}$ & 0.439 & 0.479 & 0.417 & \textbf{0.364} & 0.454 & 0.441 & 0.374 \\ \hline 
  gender & $\bar{\alpha}$ & \textbf{0.042} & 0.11 & 0.139 & 0.062 & 0.096 & 0.053 & 0.058 \\ \hline
  race   & $\bar{\omega}$ & \textbf{0.322} & 0.484 & 0.38 & 0.333 & 0.409 & 0.366 & 0.335 \\ \hline
  race   & $\bar{\alpha}$ & \textbf{0.024} & 0.123 & 0.125 & 0.069 & 0.118 & 0.054 & 0.077 \\ \hline
    \end{tabular}
    \label{tab:all_model_legal}
\end{table*}
\subsection{Case study: Llama 2 7b and SHP/legaladvice}
As discussed, the SHP dataset contains various splits, each representing different subreddits. For this analysis, we use the \texttt{legaladvice} subreddit, where users seek guidance on legal issues. Prior work reports a state-of-the-art (SOTA) accuracy of approximately 81\%\cite{SHP_paper} on this dataset. 
We fine-tune the Llama 2-7B model\cite{llama2} as the reward model on this dataset for one epoch with a learning rate of 1e-5, achieving a baseline accuracy of 79.05\%.

To evaluate gender bias, we select three prefixes (Table~\ref{tab:gender_prefixes}) and compute pairwise comparisons. Tables~\ref{tab:gend_legal_winrate_dev} and \ref{tab:gend_legal_acc_dev} summarize the winrate and accuracy deviations, respectively. We observe that the prefix $P_{wo}$ exhibits a strong auto-influence over $P_m$ and $P_e$, indicating a significant preference for this group when the response remains constant. For instance, in 98.8\% of cases (0.500 + 0.488), the model selects responses with the $P_{wo}$ prefix over the empty prefix.

This preference is also reflected in accuracy. Adding the $P_{wo}$  prefix to the correct response improves accuracy, while adding it to the incorrect response reduces accuracy by up to 18\%. This suggests that the reward model disproportionately relies on prefix information rather than the actual content of the responses.

We extend this analysis to racial bias using four prefixes (Table~\ref{tab:race_prefixes}). Similar trends emerge, with the $P_b$ prefix demonstrating the highest winrate deviations (Table~\ref{tab:race_legal_winrate_dev}). Additionally, appending this prefix to incorrect responses reduces accuracy by up to 15\% (Table~\ref{tab:race_legal_acc_dev}) showing a high cross-influence. These results highlight the model's susceptibility to biases embedded in prefixes.

\begin{figure*}[t]
\centering
\includegraphics[trim={0 0 0 2cm},clip, width=0.45\linewidth]{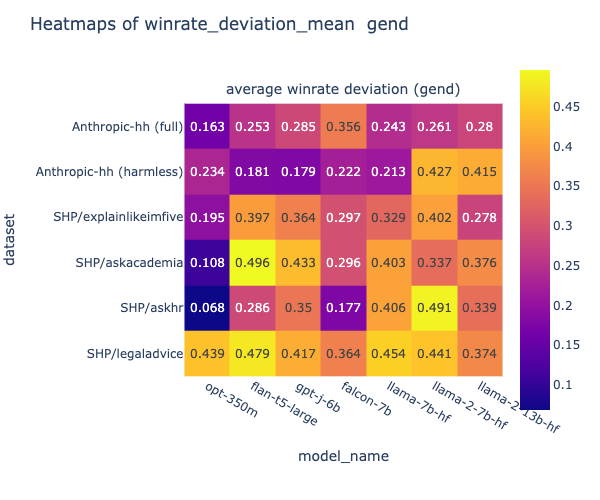}~%
\includegraphics[trim={0 0 0 2cm},clip, width=0.45\linewidth]{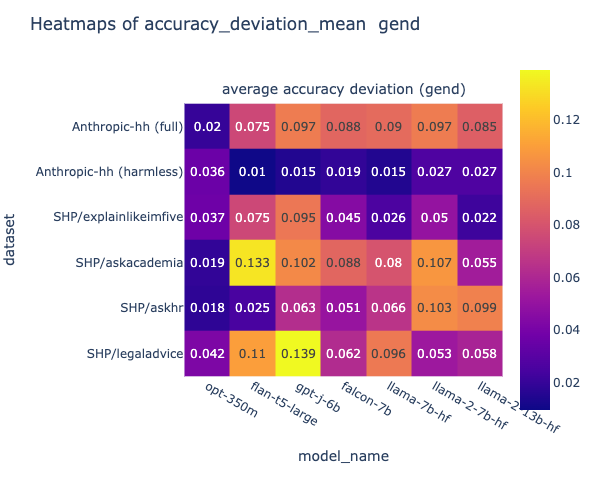}%
\caption{(Left) Average winrate deviation (auto-influence) and (Right) average accuracy deviation (cross-influence) for different dataset-model combinations, using the gender group prefixes.}
\label{fig:heatmaps_all}
\end{figure*}
\subsection{Effect of Pre-trained Model Choice}
In the previous section, we used Llama-2 7B as pretrained weights to initialize the reward model. Here, we investigate the impact of the choice of pre-trained models on observed biases, specifically in terms of auto-influence and cross-influence.

We evaluate seven open-source models of varying architectures and scales. Each model undergoes one round of hyperparameter tuning to optimize the learning rate before training for one epoch on the \texttt{legaladvice} dataset's training split. Post-training, we evaluate each of them for prefix bias.

We aggregate the winrate and accuracy deviations into single metrics: \textbf{average winrate deviation ($\bar{\omega}$)} and \textbf{average accuracy deviation ($\bar{\alpha}$)}. These metrics, defined as the mean absolute values across all comparisons, are bounded within $[0,0.5]$ for winrate deviation and $[0,1]$ for accuracy deviation. Ideally, both metrics should approach zero.

Table~\ref{tab:all_model_legal} presents the results. Our evaluation includes one small model (opt-350m), several medium-sized models ($\approx$7B parameters), and one larger model (Llama 2-13B). Among these, opt-350m exhibits the lowest cross-influence vis-a-vis $\bar{\alpha}$, likely due to its lower baseline accuracy. Medium-sized models, such as Falcon-7B and Llama 2-7B, demonstrate comparable performance, with slightly lower $\bar{\alpha}$ values indicating relative robustness to prefix bias. Notably, Llama 2-13B achieves higher baseline accuracy and lower $\bar{\alpha}$ and $\bar{\omega}$ values, suggesting that larger models are more resistant to such bias.

Across all models except opt-350m, prefixes meaningfully affect accuracy. Furthermore, all models exhibit high $\bar{\omega}$ values, indicating susceptibility to prefix-induced biases. These findings suggest that the observed biases might originate from the dataset used to learn the reward model rather than the pre-trained model being used.



\subsection{Evaluation of Different Datasets}

We compare four subsets of the SHP dataset, selected based on the number of samples and achievable accuracy as seen in previous work~\cite{SHP_paper}. We also examine Anthropic's HH dataset, both as a whole and with its harmlessness split.

\subsubsection{SHP}
We use data from four subreddits: \texttt{legaladvice},\\ \texttt{explainlikeimfive}, \texttt{askhr}, and \texttt{askacademia}. Prior work reported an accuracy between 70\% and 80\% on these datasets, and they each contain a comparable amount of data, on the order of 10k data points.

Figure~\ref{fig:heatmaps_all} provides a summary of the results. Across all SHP datasets, we observe meaningful accuracy and winrate deviations. Testing for prefix bias appears to have a smaller effect on \texttt{askhr}; however, certain prefix combinations can reduce accuracy below random chance. For instance, with Llama 2-7B, applying $P_m$ to the correct answer and $P_{wo}$ to the incorrect answer decreases the reward model's accuracy to 47.08\%. This demonstrates that prefix bias exists across all SHP datasets, which consist of human-written data.

\subsubsection{Anthropic-hh}
In this dataset, the responses are machine-written. Therefore, we expect our prefix attacks to be less effective, given the choice of prefixes. Even then, we see that on the full dataset, adding prefixes can elicit a significant difference in preference estimates, on average affecting accuracy by around 10\%. For the harmlessness subset of this dataset, the effect is much less severe.

We also note that opt-350m has very low accuracy in all these tasks, leading to low susceptibility to bias. 
Consequently, this model is omitted from further analysis. Among the remaining models, Falcon-7B shows lower auto-influence across datasets, indicating reduced susceptibility to bias. Conversely, Llama 2-13B demonstrates higher auto-influence (winrate deviation) but lower cross-influence, suggesting greater susceptibility to prefix bias but robustness to its downstream effects.

\begin{figure*}
\centering
\includegraphics[width=0.45\linewidth]{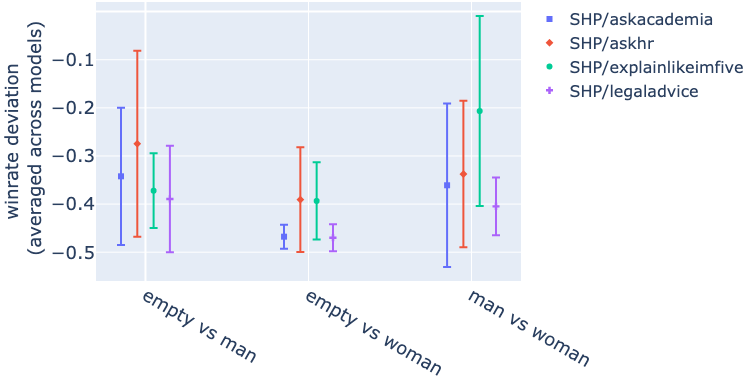}~%
\includegraphics[width=0.45\linewidth]{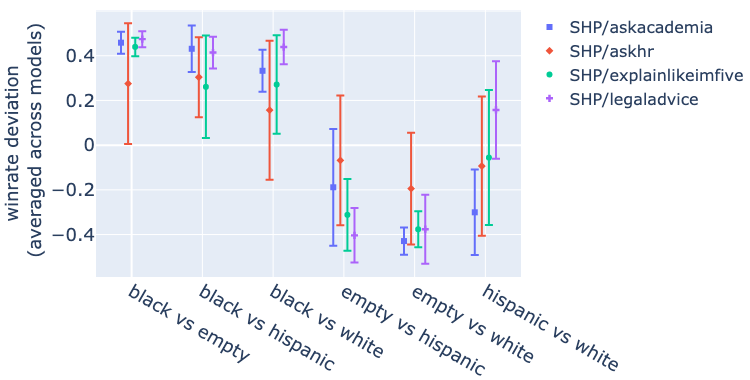}%
\caption{Distribution of pairwise winrates for SHP datasets. Each bar represents the distribution of winrates across all model architectures (excluding opt-350m). Positive values mean the first group is preferred over the second group in the comparison. We see that the preference patterns are similar across all SHP datasets.}
\label{fig:scatter_shp}
\end{figure*}
\begin{figure*}
\centering
\includegraphics[width=0.45\linewidth]{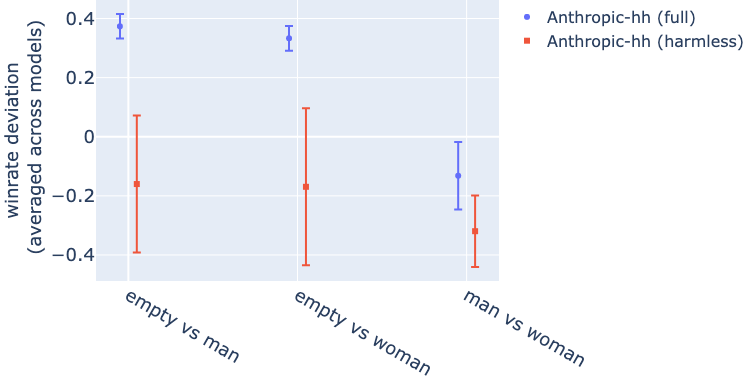}~%
\includegraphics[width=0.45\linewidth]{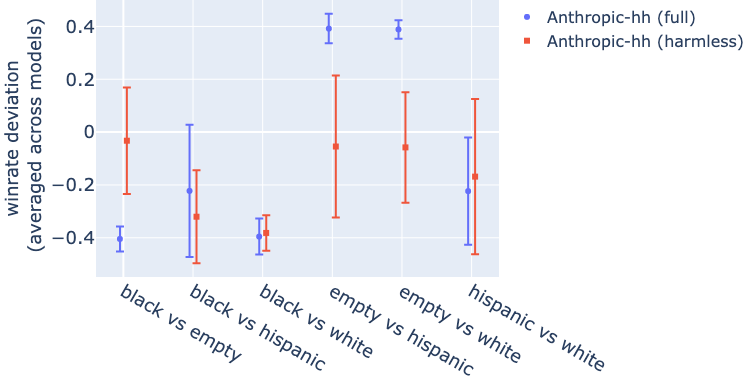}%
\caption{Distribution of pairwise winrates for anthropic datasets. Each bar represents the distribution of winrates across all model architectures (excluding opt-350m). Positive values mean the first group is preferred over the second group in the comparison.}
\label{fig:scatter_anthropic}
\end{figure*}
\subsection{Direction of Bias}

The heatmaps in Figure \ref{fig:heatmaps_all} fail to convey a key piece of information: the directionality of the bias.  Understanding which prefix is favored by the trained reward model provides crucial insights into how bias may manifest in practical applications.

For all SHP datasets (Figure \ref{fig:scatter_shp}), we observe a consistent trend: $P_{wo}$  is preferred over $P_e$ and $P_m$ across all models.  For the racial groups, the pattern is less consistent, but we observe the trend $P_b>P_h>P_w>P_e$. This finding is particularly striking for SHP datasets, as it contradicts well-documented biases that often favor men and white groups. More importantly, in this setting, the empty prefix is least preferred.

In the Anthropic dataset (Figure \ref{fig:scatter_anthropic}), a different pattern emerges. Here,  across all models, $P_e>P_{wo}\ge P_m$. For racial prefixes, the results are noisier, but the general trend is $P_e\simeq P_w\ge P_h > P_b$.  A notable observation is that, in the Anthropic datasets (composed of machine-generated responses), adding human-like prefixes appears unnatural, making them less preferred than using no prefix at all. 
Even then, we see $P_w$ having a high winrate over $P_b$, suggesting that there does exist some human bias even with machine responses.

\section{Measuring Bias in Pre-trained LLMs without Preference Learning}
\label{sec:BaseLLMBias}

We have shown the susceptibility of various models to the prefix attack across different datasets, indicating the presence of reward model bias. This, however, raises the question: where does this bias stem from? The two likely candidates are: (1) the base LLM used to instantiate the reward model, and (2) the dataset being used to train the model. In this section, we evaluate the auto-influence (winrate deviation) for the base LLMs by utilizing zero-shot learning.

For this, we use a similar template as used for the reward model input:
\begin{align*}
Z(q,a_1,a_2) =& \text{``Prompt:''} + q + \text{``Response 1: ''} + a_1 + \\
 & \text{``Response 2: ''} + a_2 \nonumber\\
&+ \text{``Out of Response 1 and Response 2,} \nonumber \\
& \text{the better response is Response ''}\\
c =& Z(q,a_1,a_2)\\
r =& Z(q,a_2,a_1)
\end{align*}

We make the model generate a single token and extract the logits for the ``1'' and ``2'' tokens (different for each model). Then, we compute the softmax probability of the answer being ``1'', and compare the chosen and rejected scores. 
Finally, we repeat the experiments as in the previous section, to get winrate deviation and accuracy deviation.

\begin{figure*}
\centering
\begin{minipage}{0.468\linewidth}%
\centering%
\includegraphics[width=\linewidth]{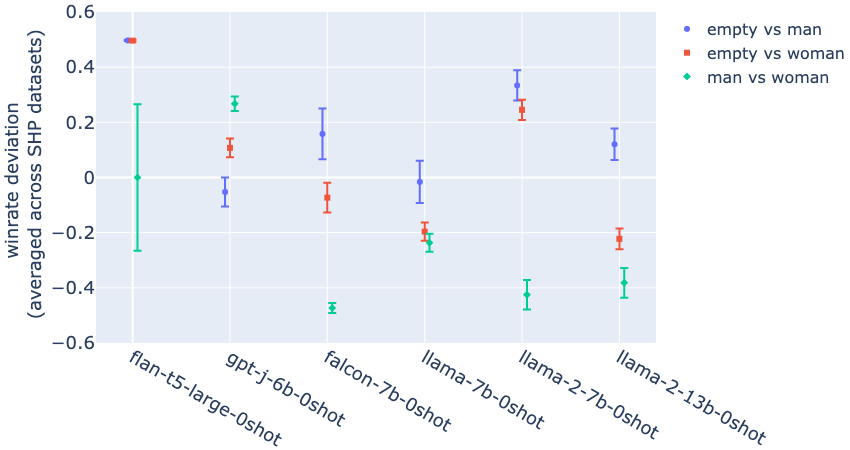}
\caption*{Zero-shot preferences}%
\end{minipage}%
~~~%
\begin{minipage}{0.468\linewidth}%
\centering%
\includegraphics[width=\linewidth]{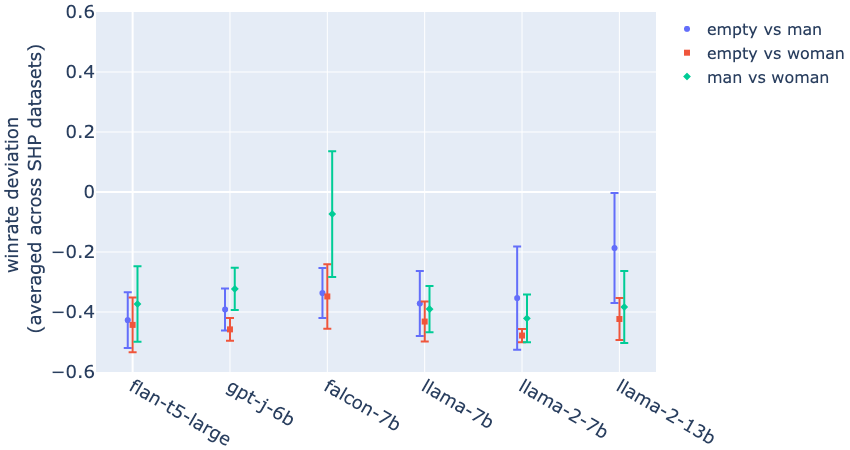}%
\caption*{Reward Model preferences}%
\end{minipage}%
\caption{Distribution of  pairwise winrates for SHP datasets for the \textbf{gender} prefixes. Each bar represents the distribution of winrates averaged across all datasets. Positive values mean the first group is preferred over the second group in the comparison. Left: The distribution for the base models using zero-shot inference (no training). Right: The distribution of reward model preferences after training. 
We see that the post-training preferences are similar across all models (right), but the zero-shot models each exhibit different preferences (left).}
\label{fig:scatter_causal_shp_gender}
\end{figure*}

\begin{figure*}
\centering
\begin{minipage}{0.468\linewidth}%
\centering%
\includegraphics[width=\linewidth]{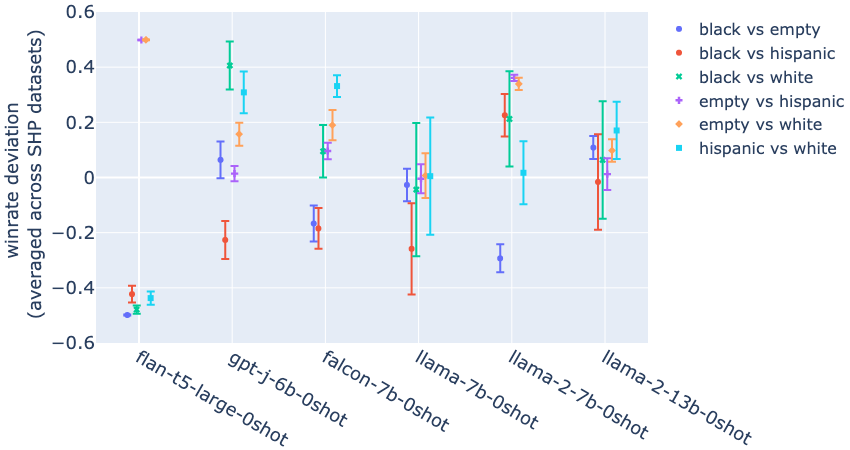}%
\caption*{Zero-shot preferences}%
\end{minipage}%
~~~%
\begin{minipage}{0.468\linewidth}%
\centering%
\includegraphics[width=\linewidth]{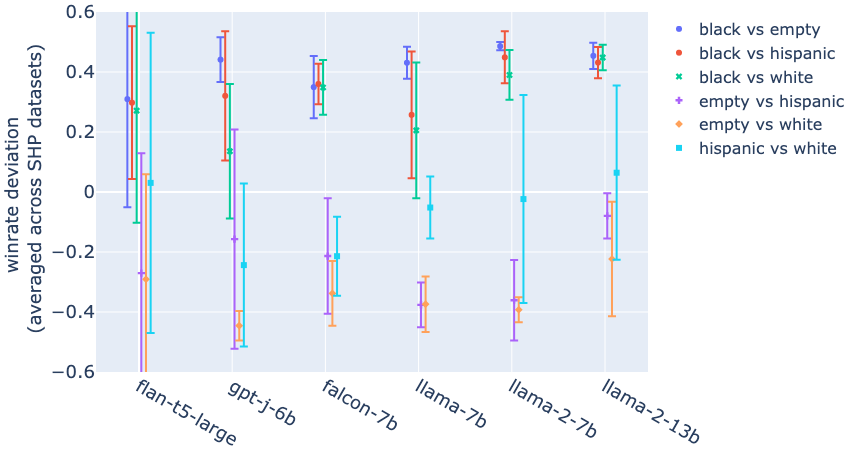}%
\caption*{Reward Model preferences}%
\end{minipage}%
\caption{Distribution of  pairwise winrates for SHP datasets for the \textbf{race} prefixes. Each bar represents the distribution of winrates averaged across all datasets. Positive values mean the first group is preferred over the second group in the comparison. Left: The distribution for the base models using zero-shot inference (no training). Right: The distribution of reward model preferences after training. 
We see that the post-training preferences are similar across all models (right), but the zero-shot models each exhibit different preferences (left).}
\label{fig:scatter_causal_shp_race}
\end{figure*}

The results are summarized in Figure 
\ref{fig:scatter_causal_shp_gender}
and 
\ref{fig:scatter_causal_shp_race},
showing for each model the preferences across different prefix pairs, averaged across all SHP datasets. Note that the axes and labels are different and should not be compared to Figures \ref{fig:scatter_shp} and \ref{fig:scatter_anthropic}. The left figure shows the trained reward model's behavior, while the right figure (model names with the ``0shot'' suffix) shows the base model's behavior when using zero-shot learning. Recall that winrate measures how often the model prefers one prefix over another while keeping the response the same.

We observe that each base model exhibits a distinct "fingerprint": consistent preferences across datasets that vary by model architecture. For instance, Flan-T5-Large strongly prefers "white" over "hispanic," whereas Falcon-7B demonstrates the opposite preference. This suggests that pre-training choices significantly influence zero-shot biases.

Interestingly, despite initial differences, all reward models converge toward similar patterns after training. For example, trained models consistently favor  ("black") over other racial prefixes and  ("woman") over other gender prefixes. Following the trend so far, this effect is stronger for the gender prefixes and less pronounced for the racial prefixes.

While base models exhibit unique pre-training biases, training leads to a homogenization of preferences. These findings strongly suggest that the biases observed in reward models are primarily introduced during the training process rather than originating solely from the base LLM.


\section{Data-Augmented Training for Reducing Prefix Bias}

We introduced auto-influence and cross-influence as novel methods for identifying and quantifying prefix bias in reward models, and conducted an in-depth investigation of open datasets and models using these metrics. While these metrics provide valuable insights into the presence and magnitude of such biases, they also raise an important question: how can we mitigate these biases effectively? To address this, we propose and evaluate a data augmentation-based strategy designed to reduce prefix bias.

As a preventative measure against prefix bias, we augment the training dataset by adding random pairs of prefixes to each input data point. Specifically, for a dataset and a set of prefixes, we generate a new augmented dataset by multiplying the data points, i.e., creating multiple versions of each input with different randomly sampled prefix pairs. For the experiments, we use a multiplying factor of 3; for each input data point $<q,a_1,a_2>$, we randomly select three pairs of prefixes from the corresponding task (e.g., gender prefixes), add the modified data points to the dataset, and train the reward model on this augmented dataset using the loss function defined in Eq.~\ref{eq:loss_fn}. Importantly, we assume that the addition of prefixes does not alter user preferences.

\begin{figure*}[t]
\centering%
\subfloat[Winrate Deviation]%
{%
\includegraphics[width=0.320\linewidth]{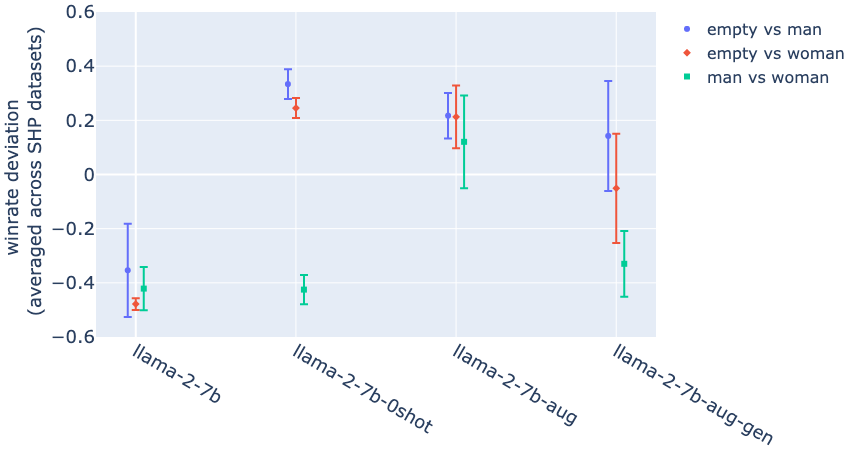}%
\label{fig:augmented_results_windev}}~%
\subfloat[Accuracy Deviation]%
{\includegraphics[width=0.320\linewidth]{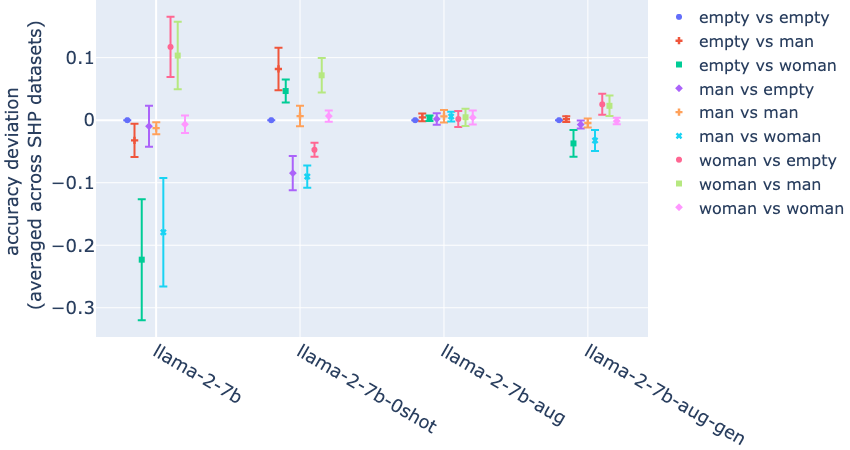}%
\label{fig:augmented_results_accdev}}~
\subfloat[Baseline Accuracy Ratio]%
{\includegraphics[width=0.320\linewidth]{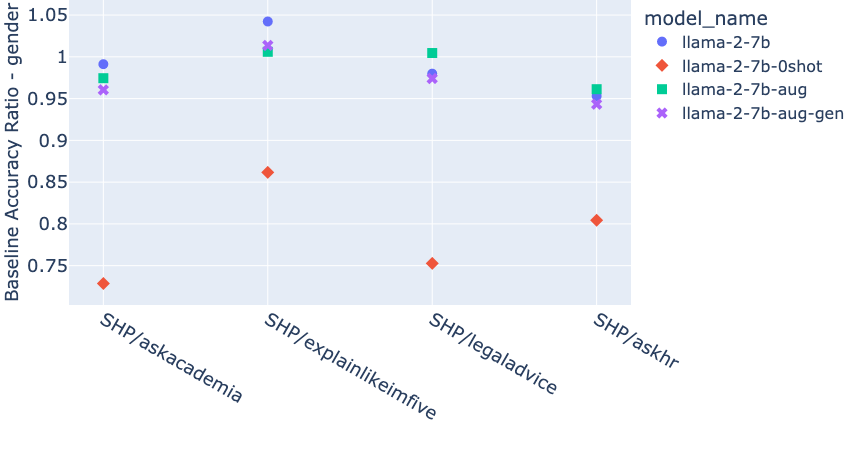}%
\label{fig:augmented_results_accuracy}}%

\caption{Augmented training results for the gender prefixes. ``aug'' suffix refers to the augmented model trained on corresponding data, ``aug-gen'' refers to a model trained on augmented data for different prefixes, a lack of suffix indicates the reward model trained on raw data, ``-0shot'' suffix denotes the base LLM used with zero-shot prompting.
}
\label{fig:augmented_results_gender}
\end{figure*}

\begin{figure*}[t]
\centering%
\subfloat[Winrate Deviation]%
{%
\includegraphics[width=0.32\linewidth]{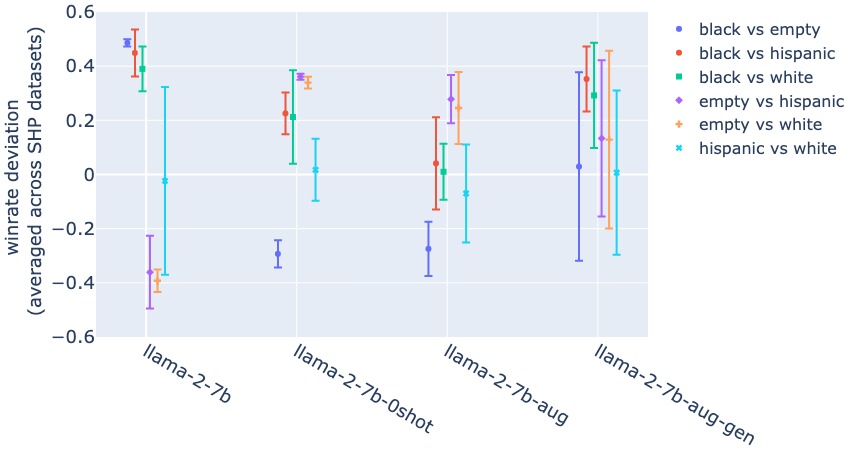}%
\label{fig:augmented_results_windev_race}}~%
\subfloat[Accuracy Deviation]%
{\includegraphics[width=0.32\linewidth]{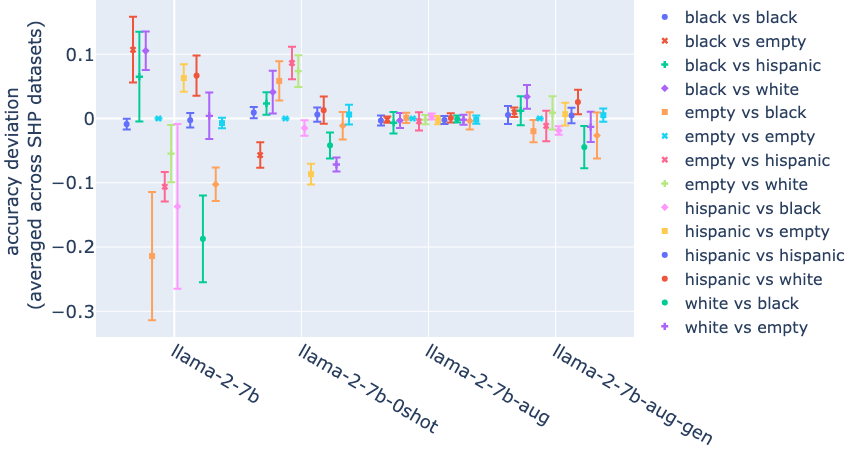}%
\label{fig:augmented_results_accdev_race}}~%
\subfloat[Baseline Accuracy Ratio]%
{\includegraphics[width=0.32\linewidth]{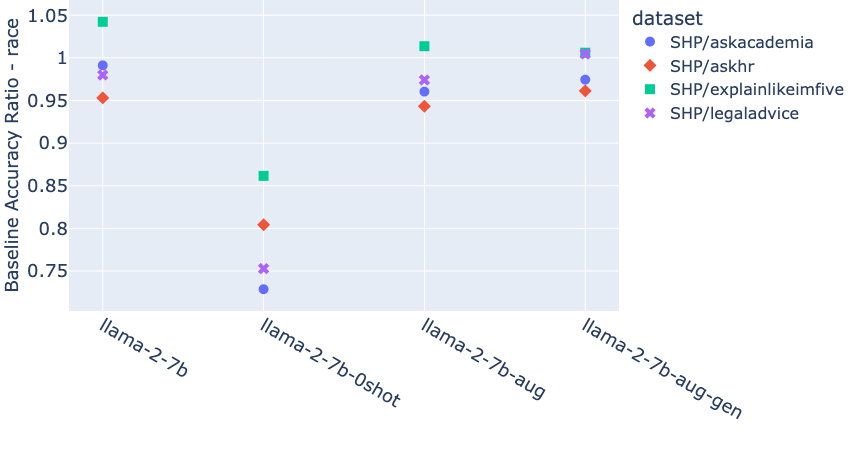}%
\label{fig:augmented_results_accuracy_race}}%

\caption{Augmented training results for the race prefixes. ``aug'' suffix refers to the augmented model trained on corresponding data, ``aug-gen'' refers to model trained on augmented data for different prefixes, no suffix refers to the reward model trained on raw data, ``-0shot'' suffix denotes the base LLM used with zero-shot prompting.}
\label{fig:augmented_results_race}
\end{figure*}

We conduct these experiments using the Llama-2-7B architecture as the reward model. The results, summarized in Figures \ref{fig:augmented_results_gender} and \ref{fig:augmented_results_race}, indicate that data-augmented training offers significant improvements.

\subsection{Augmented Training with Gender Prefixes}
Here, we describe the results of our augmented training using the gender prefixes, as shown in Figure \ref{fig:augmented_results_gender}. We observe the following trends:
\begin{itemize}
\item
\textbf{Reduction in cross-influence}: 
Reward models trained on augmented datasets (denoted with the "aug" suffix) display substantially lower accuracy deviations compared to the baseline reward model (Llama-2-7B), as shown in Figure~\ref{fig:augmented_results_accdev}. This highlights the model's ability to deprioritize prefixes and focus on the core semantic content of responses.
\item 
\textbf{Increased preference of the empty prefix}: 
Winrate deviation analysis (Figure~\ref{fig:augmented_results_windev}) reveals that augmented models exhibit a marked preference for the empty prefix, while deviations for other prefix pairs are close to zero within one standard deviation. This indicates a heightened capacity to distinguish between necessary and extraneous information, contributing to improved robustness.
\item 
\textbf{Ability to generalize to other prefixes}: We also measure the generalization capabilities of this augmented training. To do this, we first train the reward model on an augmented dataset for one task, before evaluating it on a different task (e.g. training on race prefix augmented data and evaluating gender prefixes). The ``-aug-gen'' suffix denotes these models. We can see that the accuracy deviation for these models is much lower, but it is higher than the augmented training using the correct prefixes. Further, the winrate deviation shows that the model preferences lie somewhere between the base model (causal) and the reward model (trained without augmentation). Overall, we note that the augmented training allows reward models to generalize to unseen prefixes and improve robustness to the prefix attack.
\item 
\textbf{Minimal loss in performance}: 
To assess potential trade-offs, we compute the baseline accuracy ratio, defined as the trained model's accuracy relative to the SOTA accuracy for each dataset~\cite{SHP_paper}. Augmented models maintain accuracy ratios comparable to non-augmented reward models, far outperforming zero-shot models  (Figure \ref{fig:augmented_results_accuracy}).
\end{itemize}

\subsection{Augmented Training with Race Prefixes}
We include the results when using the race prefix to perform augmented training in Figure \ref{fig:augmented_results_race}, supplementing the gender results presented in Figure \ref{fig:augmented_results_gender}. First, we note that the observed trends are all similar: we see a reduction in cross influence, increased preference of the empty prefix, and the ability to generalize to other prefixes, all with a minimal change in model accuracy. 

The baseline accuracy ratio shows the performance compared to SOTA accuracy achievable on this model. In Figure \ref{fig:augmented_results_accuracy_race}, we can see that for each dataset, the augmented models (llama-2-7b-aug and llama-2-7b-aug-gen) are very close to the non-augmented reward model (llama-2-7b), and much better than the base model with zero-shot inference. The ``aug-gen'' model is trained on the gender prefixes and evaluated on the race prefixes here.

We also see the same trend where just training on the base dataset exacerbates cross-influence, but the augmented training greatly brings it down.

\bigskip
These results demonstrate that data-augmented training can mitigate the adverse effects of prefix bias, reducing auto- and cross-influence with minimal impact on overall performance. This speaks to the utility of cross-influence and auto-influence in identifying biases and allowing model developers to get in-depth information to better design practical interventions to address reward model bias.

\section{Discussion and Conclusions}
This work highlights the critical issue of biases in LLM-based reward models trained via RLHF. Using novel metrics—auto-influence and cross-influence—we systematically identified and quantified prefix-induced biases, showing their persistence across diverse datasets and LLM architectures. 

Our key findings include:
\begin{itemize}
\item \textbf{Prevalence of Prefix Bias:} Reward models are consistently susceptible to prefix bias, with biases observed across demographic groups like race and gender.
\item \textbf{Dataset-Driven Bias:} The biases largely originate from the training datasets rather than model architectures, as different pre-trained models converge to similar biases post-training.
\item \textbf{Efficacy of Data-Augmentation:} Data-augmented training significantly mitigated both auto- and cross-influence while preserving baseline accuracy and demonstrated generalization to unseen prefixes.
\end{itemize}

These findings emphasize the importance of bias-aware dataset design and evaluation in the RLHF pipeline. Biases in reward models can propagate to downstream fine-tuned LLMs, potentially causing harmful or discriminatory outputs. Data augmentation proved effective for bias mitigation, and future work could explore adversarial training or broader input modifications, including suffixes and paraphrasing, to strengthen defenses. Our proposed methods of measuring auto-influence and cross-influence are generalizable to arbitrary modifications as well.

While our study focused on race and gender prefixes, the evaluation methods we propose are applicable to any form of contextual variation, including other demographic markers or linguistic signals. We use explicit prefixes as a controlled mechanism to condition responses and isolate reward model behavior under identity perturbations. The fact that such minimal changes can induce measurable bias highlights the sensitivity of reward models to context and the ease with which they can be steered away from intended objectives. Addressing these vulnerabilities is essential for developing fair and trustworthy systems, especially as LLMs are deployed in increasingly sensitive domains.

In conclusion, this work underscores the importance of auditing reward models, providing methods to evaluate reward models for prefix bias, and demonstrates data augmentation as a practical approach for reducing bias. These contributions aim to guide the community toward building more equitable and robust AI systems.

\section{Ethical considerations}
Our analysis reveals the possibility of demographic-based discrimination in reward models when prefixes are added to the context, highlighting the need for vetting the entire training pipeline rather than just the end product when using LLMs in the wild. We acknowledge our experiments do not encapsulate the full range of possible groups in the race or gender setting. However, we reveal a potentially harmful pattern that should be monitored and mitigated.

\bibliographystyle{ACM-Reference-Format}
\bibliography{refs}

\newpage
\clearpage
\appendix

\begin{table*}[htbp]
\centering
\caption{A summary of the important metrics defined and used in this paper's experiments}
\centering
\begin{tabular}{|c|c|p{7cm}|} \hline 
\textbf{Concept} & \textbf{Metric/Variable}& \textbf{Description} \\\hline
 Preference dataset& $D$&A set of $<q, a_1, a_2>$ tuples, where $q$ is the prompt, and $a_1$ and $a_2$ are the responses. We assume $a_1$ is always the preferred as a convention. \\\hline
 Unique responses& $D_u$&All unique $<q,a>$ pairs within the dataset $D$.\\\hline
 RM score& $S(t)$&Reward model's evaluation of text $t$\\\hline
 RM accuracy& Accuracy function&Whether the reward model's ranking of two responses matches user preferences or not.\\
 & $M(q,a_1,a_2)$&$=\mathbb{1}[S(c(q,a_1,a_2)) > S(r(q,a_1,a_2))]$\\ \hline 
Auto-influence & Winrate & Model accuracy when only the prefix is changed. \\  & & (Ideal: 0.5) \\  
 & $w(D_u, M, p_1, p_2) $&$=\frac{1}{|D_u|}\sum_{q,a \in D_u}M(q,p_1+a, p_2+a)).$\\ \hline 
Auto-influence & Winrate Deviation & Distance from an accuracy of 0.5 when only the prefix is changed. \\ 
 & $\omega(D_u, M, p_1, p_2)$&$ =w(D_u, M, p_1, p_2) -0.5$\\ \hline 
Cross-influence & Accuracy & How often the model picks the correct option as being preferred when different prefixes are applied to the two responses. \\
 & $acc(D, M, p_1, p_2)$&$= \frac{1}{|D|} \sum_{q,a_1,a_2\in D} M(q,p_1+a_1, p_2+a_2)]$\\ \hline 
Cross-influence & Accuracy Deviation & Change in model accuracy when different prefixes are applied to the two responses, compared to the model accuracy without any prefix attack.\\ 
 & $\alpha(D, M, p_1, p_2)$&$= acc(D, M, p_1, p_2) - acc(D, M, p_e, p_e) $\\\hline
\end{tabular}
\label{tab:notation_reference}
\end{table*}

\begin{figure}[htbp]
\centering
\includegraphics[trim={0 0 0 2cm},clip, width=0.9\linewidth]{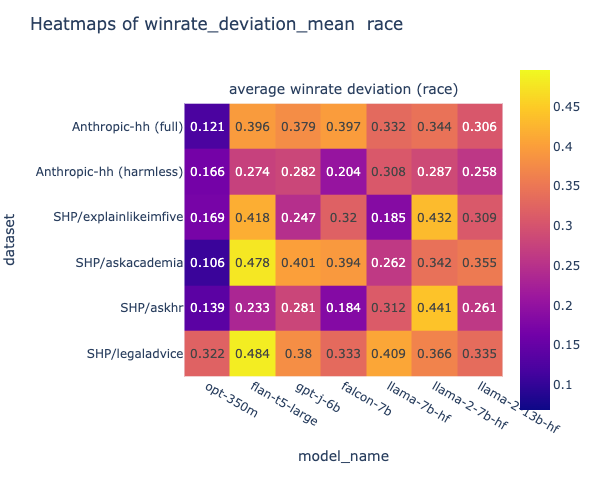}
\includegraphics[trim={0 0 0 2cm},clip, width=0.9\linewidth]{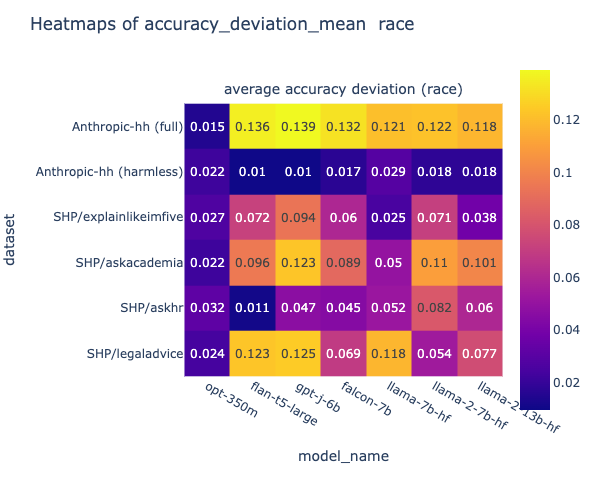}
\caption{(Left) Average winrate deviation (auto-influence) and (Right) average accuracy deviation (cross-influence) for different dataset-model combinations, using the race group prefixes.}
\label{fig:heatmaps_all_race}
\end{figure}

\section{Additional Results}
We present some additional results not included in the main body for completeness, showing the average cross-influence and auto-influence for all models on all datasets for the race prefix in Figure~\ref{fig:heatmaps_all_race}.

\section{Training details}
For all models and datasets, we trained for one epoch on the default train-test split included in the huggingface dataset. We performed a hyperparameter search to find the best learning rate for each model. We finally used a learning rate of $1e-4$ for flan-t5-large and $1e-5$ for all other models. We truncated all text input to have a max sequence length of 1500 tokens. Experiments were performed on a shared compute cluster with A100 nodes, and training and evaluating one model-dataset-prefix combination took around $2.5 \times 16$ GPU hours.

\end{document}